\pdfoutput=1
\documentclass{article}
\usepackage{amsmath}
\usepackage{spconf,graphicx}
\usepackage{amsfonts} 
\usepackage{amssymb}
\usepackage{amsthm}
\usepackage{multirow}
\usepackage{multicol}
\usepackage{algorithm}
\usepackage[noend]{algpseudocode}
\usepackage{enumerate}
\usepackage{subcaption, caption}
\usepackage[flushleft]{threeparttable}
\usepackage{float}
\usepackage{xspace}
\usepackage{caption}
\usepackage{diagbox}
\usepackage{tabularx}
\usepackage{stfloats}
\usepackage{graphicx}
\usepackage{amsthm}
\usepackage{enumitem}

\setlist[enumerate]{topsep=0pt, partopsep=0pt, parsep=0pt, itemsep=0pt}
\setlist[itemize]{topsep=0pt, partopsep=0pt, parsep=0pt, itemsep=0pt}
\usepackage{adjustbox}
\usepackage{booktabs}
\usepackage[numbers,sort&compress]{natbib}

\def\methname{SecureCut}
\title{SecureCut: Federated Gradient Boosting Decision Trees with Efficient Machine Unlearning}
\name{Jian Zhang$^1$, Bowen Li $^1$, Jie Li$^{1,2,*}$, Chentao Wu$^1$}
\address{$^1$Department of Computer Science and Engineering, Shanghai Jiao Tong University, China\\
$^2$MoE Key Lab of Artificial Intelligence, AI Institute,
Shanghai Jiao Tong University, China\\
}
\begin{document}
%\ninept
%
\maketitle
\begin{abstract}
% With legislation mandating companies to honor the 'right to be forgotten' by erasing user data, addressing removal requests in machine learning has become imperative.
% To address this issue, we explore the problem of machine unlearning in Vertical Federated learning(VFL).
% In VFL, where multiple parties provide distinct features for training, erasing data for privacy often necessitates removing specific features across all samples.
% In this paper, we propose a novel GBDT-based model that effectively achieves both instance unlearning and feature unlearning without retraining from scratch.
% In particular, while constructing the tree, we consider the robustness of its structure during the unlearning phase and stored intermediate results to minimize computational efforts in the unlearning process.
% Extensive experiments results on 3 datasets demonstrate that our model  achieves superior performance relative to baseline models. 
% To our best knowledge, this is the first work that investigate machine unlearning in VFL.
In response to legislation mandating companies to honor the \textit{right to be forgotten} by erasing user data, it has become imperative to enable data removal 
% in privacy-aware machine learning, especially 
in Vertical Federated Learning (VFL) where multiple parties provide private features for model training. 
In VFL, data removal, i.e., \textit{machine unlearning}, often requires removing specific features across all samples under privacy guarentee in federated learning. To address this challenge, we propose \methname, a novel Gradient Boosting Decision Tree (GBDT) framework that effectively enables both \textit{instance unlearning} and \textit{feature unlearning} without the need for retraining from scratch. Leveraging a robust GBDT structure, we enable effective data deletion while reducing degradation of model performance. Extensive experimental results on popular datasets demonstrate that our method achieves superior model utility and forgetfulness compared to \textit{state-of-the-art} methods. To our best knowledge, this is the first work that investigates machine unlearning in VFL scenarios.
\end{abstract}
\section{Introduction}
\vspace{-10pt}

Machine learning models trained with tremendous amount of data memorize the training data \cite{carlini2019secret, carlini2021extracting}, however recently enforced legislation requires companies to delete users’ data records upon privacy request \cite{regulation2018general}. Therefore, is an essential capability of the models to delete specific data records, i.e., \textit{machine unlearning}. The vanilla approach for machine unlearning is to retrain a model with the remained data\cite{cao2015towards, bourtoule2021machine}, which often incurs significant costs. Addressing this problem, machine unlearning techniques are tailored for Deep Neural Networks (DNN) with efficient data removal\cite{cao2015towards, bourtoule2021machine, guo2020certified, wu2020deltagrad, gupta2021adaptive, golatkar2021mixed, neel2021descent}.

\begin{figure}[tb]
\vspace{-2pt}
\centering
\includegraphics[width=0.5\textwidth]{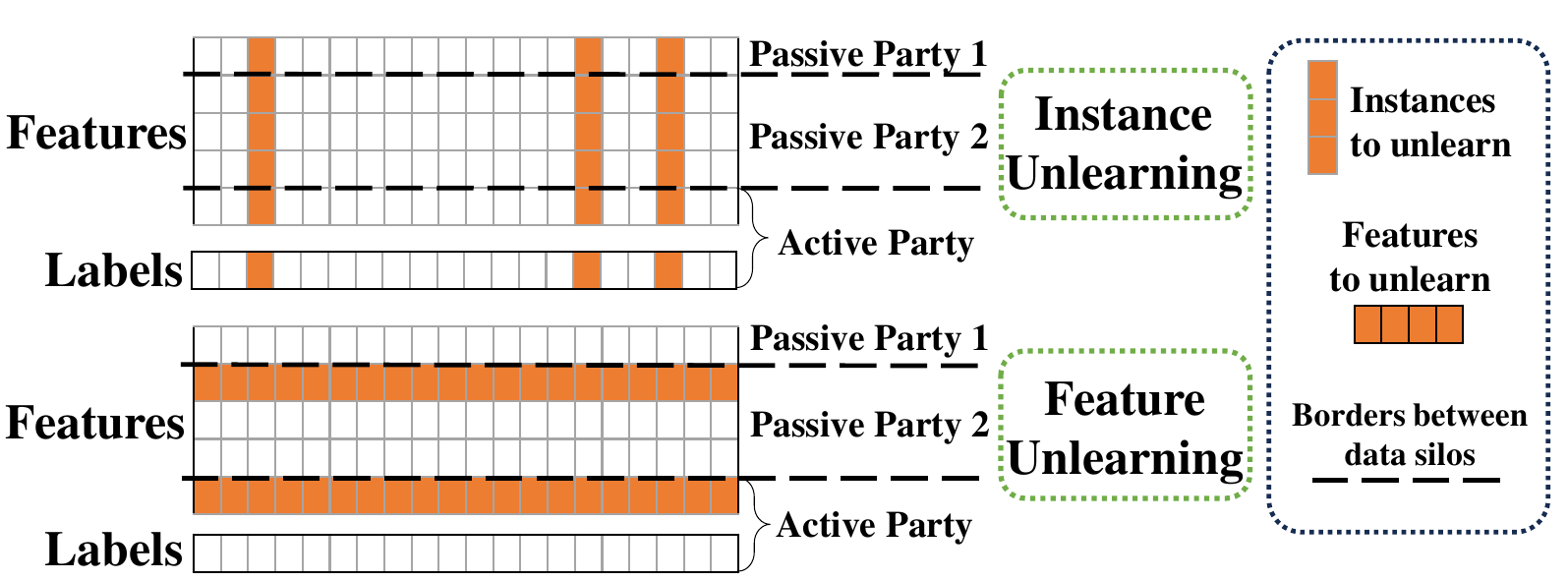}
\vspace{-15pt}
\caption{ The difference between Instance Unlearning and Feature Unlearning in VFL scenarios, where private date features are secretly maintained by multiple parties. Traditional unlearning primarily focuses on Instance Unlearning, which involves unlearning a specific instance. In contrast, our SecureCut framework considers both Instance Unlearning and Feature Unlearning, making it more suitable for VFL scenarios.}
\label{fig:task}
\vspace{5pt}
\end{figure}

Vertical Federated Learning (VFL) is a privacy-aware training method where multiple parties collaboratively train a model without disclosing private training data. Specifically, parties contribute different features to model training, thus the data deletion request in VFL means deleting specific features across all samples,  i.e., \textit{feature unlearning}, see Fig. \ref{fig:task}. Previous machine unlearning methods in federated learning settings only delete data instances from a trained model, i.e., \textit{instance unlearning}, but \textit{feature unlearning} remains an open problem to be solved in VFL\cite{gao2022verifi, liu2021federaser, liu2022right}. 
Moreover, \textit{Vertical Federated Tree Boosting} (VFTB) is a highly representative and widely adopted algorithm in industrial applications such as financial data mining and recommendation systems, of which secureboost is considered a top performer\cite{cheng2021secureboost,chen2021secureboost+,NEURIPS2022_84b74416,zhao2022sgboost}. In this paper, we give solutions for both \textit{instance unlearning} and \textit{feature unlearning} in  Vertical Federated Tree Boosting (VFTB) scenarios. 

There are three main technical challenges when unlearning data instances and features in VFTB models. Firstly, deleting a data record in the parent node causes a \textit{chain reaction}. This reaction leads to changes in the instance space of child nodes and affects the gradients in subsequent boosting trees\cite{wu2023deltaboost}. Secondly, VFTB utilizes secret \textit{lookup tables} for node splitting during both training and inference stages. These tables require iterative modifications when data is deleted\cite{cheng2021secureboost}. Thirdly, compared to data deletion, data feature unlearning induces more significant changes in tree structures. This is because node splitting relies entirely on features. While several studies have delved into unlearning in tree-based models \cite{schelter2021hedgecut, wu2023deltaboost}, their focus has largely been on instance unlearning within VFL settings. The intricacies of implementing deletion in a VFL context, especially concerning feature unlearning, are still largely unaddressed.

To address the aforementioned challenges, we propose SecureCut, a VFTB framework tailored for efficient machine unlearning in VFL scenarios. The cornerstone of our approach is a robust tree structure such that removing a small portion of samples has a minimal impact on the optimal split point and feature selection. Starting from this characteristic, we employ techniques like \textit{Secure Binary Tree Bucketing} and \textit{Robust Splitting} to design the SecureCut Tree, thereby guaranteeing its stability during the unlearning process. Our main contributions are summarized as follows:
\begin{itemize}[leftmargin=12pt]
\item We propose the first work to investigate machine unlearning in VFL, introducing both data instance unlearning and data feature unlearning methods.
\item We introduce SecureCut, a novel VFTB-based learning framework, crafted specifically to facilitate efficient unlearning of data instances and features in VFL. Specifically, we propose Secure Binary Tree Bucketing and Robust Splitting to build robust SecureCut Trees.
\item Extensive experimental results on three tabular datasets rigorously validate that the proposed SecureCut effectively achieves both model utility and the deletion forgetfulness. 
\end{itemize}

\begin{figure*}[tb]
\centering
\includegraphics[width=1\textwidth]{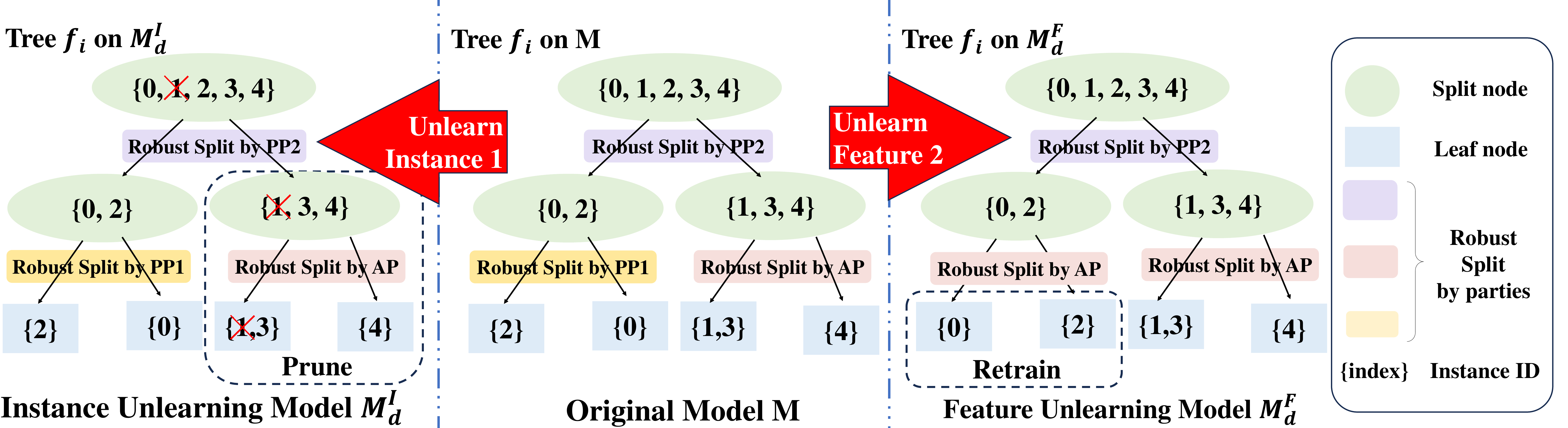}
\vspace{-10pt}

\caption{Illustration of Instance and Feature Unlearning in SecureCut Model $M$. $f_i$ is a tree within $M$, and \textit{feature 2} is provided by Passive Party 1 (PP1). Both Instance and Feature Unlearning processes can precisely unlearn the specified request.}
\label{fig:structure}
\vspace{-15pt}
\end{figure*}

\section{Problem Formulation}

\noindent\textbf{Gradient Boosting Decision Trees}:  XGBoost\cite{chen2016xgboost} is a representative GBDT model, where the model $M$ predicts output for an instance with an ensemble of $K$ regression trees: 

\begin{small}
\begin{equation}
\hat{y_i} = M(x_i) = \sum_{k=1}^{K} f_k(x_i), \quad f_k \in F, 
\label{eq:a}
\end{equation}
\end{small}
where $F$ are functions $F=\{f(x)=w_{q(x)}\}$, in which  \( q: \mathbb{R}^m \rightarrow T \) maps data instances to tree leaves and \(w \in \mathbb{R}^T \). 

Each single regression tree is recursively constructed by tree node splitting maximizing the following gain function:

\begin{small}
\vspace{-9pt}
\begin{equation}
\mathcal{G} = \frac{1}{2} \left[ \frac{\left(\sum_{i \in I_L} g_i\right)^2}{\sum_{i \in I_L} h_i+\lambda} + \frac{\left(\sum_{i \in I_R} g_i\right)^2}{\sum_{i \in I_R} h_i+\lambda} - \frac{\left(\sum_{i \in I} g_i\right)^2}{\sum_{i \in I} h_i+\lambda} \right] - \gamma,
\label{eq:G}
\end{equation}
\end{small}
where \( I_L \) and \( I_R \) represent the subsets of instances assigned to the left and right child nodes after a split. The terms \( g_i \) and \( h_i \) denote the first and second order derivatives of \( l(y_i, \hat{y_i}^{(t-1)}) \) with respect to \( \hat{y_i}^{(t-1)} \). The optimal weight \( w \) for a leaf node can be calculated as:

\vspace{-7pt}
\begin{small}
\begin{equation}
w^* = -\frac{\sum_{i \in I_j} g_i}{\sum_{i \in I_j} h_i+\lambda}
\label{eq:w}
\end{equation}
\end{small}
\vspace{-5pt}

\noindent\textbf{Vertical Federated Tree Boosting (VFTB)}: An \textit{active party} (\textbf{AP}) is the data provider with labels \( Y \in \mathbb{R}^{n \times 1} \), and we define \textit{passive parties} (\textbf{PP}) as the \( m \) data providers that only have data features without the corresponding labels. These features are denoted as $ \{ X_i \in \mathbb{R}^{n_i \times d_i} \}_{i=1}^{m}$
where \( n_i \) is the number of data samples and \( d_i \) is the data dimensionality within party \( i \). Considering parties with overlapping sets of samples, the shared data samples across parties are $ \{ X_i \in \mathbb{R}^{n \times d_i} \}_{i=1}^{m} \footnote{We use cryptographically secure algorithms to identify this intersection, which forms the training set.}. $
The maximum number of trees is defined as \( N \), and the maximum depth of a VFTB tree is given as \( H_t \).

\noindent \textbf{Unlearning Objectives in Vertical Federated Learning}: 
Consider a dataset \(D = \{(x_i, y_i)\}\) with \(n\) instances and \(m\) features, where \(x_i \in \mathbb{R}^m\) and \(y_i \in \mathbb{R}\). 

Formally, the model \(M\) is derived by \(M = \Phi_{\text{learn}}(D)\), indicating model training on dataset \(D\). For unlearning instances, \(M_d^I = \Phi_{\text{unlearn}}(M, D, D_d)\) represents the model after deleting data \(D_d\). For feature unlearning, the new model is represented as \(M_d^F = \Phi_{\text{unlearn}}(M,D,F_d)\). 

\noindent\textit{Unlearning Data Instances}:
Consider a subset of instances \(D_d\) within dataset \(D\), where \(D_d = \{(x_i, y_i)\}\) has \(n_d\) instances with \(x_i \in \mathbb{R}^m\) and \(y_i \in \mathbb{R}\). The goal is to delete \(D_d\) such that:

\vspace{-5pt}
\begin{small}
\begin{equation}
        \Phi_{\text{unlearn}}(M, D, D_d) = M_d^I \approx \Phi_{\text{learn}}(D-D_d) = M_r. 
\end{equation}
\end{small}
\vspace{-2pt}

\noindent\textit{Unlearning Data Features}:
Consider a set of feature IDs \(F_d\) from dataset \(D\), where \(F_d \subseteq \{1, 2, \dots, m\}\). The goal is to delete  \(F_d\) to approximate:

\vspace{-5pt}
\begin{small}
\begin{equation}
        \Phi_{\text{unlearn}}(M, D, F_d) = M_d^F \approx \Phi_{\text{learn}}(D/F_d) = M_r. 
\end{equation}
\end{small}
\vspace{-2pt}

\vspace{-20pt}
\section{Proposed Methods}
\vspace{-10pt}

In this section, we introduce SecureCut, an innovative model based on Gradient Boosted Decision Trees (GBDT), tailored for efficient machine unlearning within Vertical Federated Learning (VFL) frameworks. First, we delve into the training process of SecureCut, highlighting its Secure Binary Tree Bucketing and Robust Splitting mechanisms. Subsequently, we outline the processes for both instance unlearning and feature unlearning in SecureCut. Lastly, we explain the robustness of SecureCut.

\subsection{SecureCut Model Training}
\vspace{-5pt}
The training process of SecureCut comprises three main stages: 1) Model initialization 2) Secure Binary Tree Bucketing 3) Robust Splitting \& Iterative Tree Boosting

\noindent \textbf{1. Model initialization}: The initialization phase involves an active party (AP) in collaboration with \(m\) passive parties (PPs) to train a global boosting tree model over their individual datasets. Once the agreement is in place, the AP creates a set of homomorphic encryption keys, denoted as \(PK\)(public key) and \(SK\)(secret key).

\noindent \textbf{2. Secure Binary Tree Bucketing}: Each Passive Party \(i\) possesses its local training dataset, represented as \(\{\textbf{X}^i\in \mathbb{R}^{n_i\times d_i} \}_{i=1}^{m}\). Each PP then independently constructs a Binary Tree \(T_i^k\) for every feature \(k\) in \(d_i\). 

During the construction of the binary tree at node \(a\), both the maximum (\(x^k_{max}\)) and minimum (\(x^k_{min}\)) feature values are considered. If the number of instances at the node surpasses the defined maximum bucketing size \(b\), a split occurs at the median feature value. Consequently, instance bucketing is executed using all the terminal nodes. Adjustments and lookups in this tree possess a computational complexity of \(O(H_b)\), where \(H_b\) represents the height of the binary tree.

\noindent \textbf{3. Robust Splitting \& Tree Boosting}:
The global model development is a result of multiple interaction rounds. This continues until the tree count aligns with \(N\) (maximum tree number). Furthermore, AP continuously bifurcates the data instances until the depth of the \(i\)-th tree matches \(H_t\) (maximum SecureCut tree depth). This process is subdivided into:

\noindent \textit{Robust Splitting finding}: The Active Party (AP) endeavors to identify the \(t\)-th robust split (where \(t\) ranges from 1 to \(2^{H_t}-1\)). The introduced \textit{Splitting Feature Neighborhood} \(S^k_t\) signifies a cluster of nearby split candidates within a feature. Additionally, the \textit{Robust Features} \(F_t\) are those features for which the difference between \(\mathcal{G}_f\) and \(\mathcal{G}_{max}\) is less than \(\epsilon\) (a predefined hyper-parameter). The detailed process is descibed in Algorithm \ref{alg:split}.

\noindent \textit{Tree Boosting}: In this phase, the model computes the weights of every leaf node based on Equation \ref{eq:w} and augments the tree based on Equation \ref{eq:a}.

\begin{algorithm}[tb]
\caption{SecureCut Robust Splitting Finding}
\begin{algorithmic}[1]
    \State /* AP: active party, PP: passive party */
    % \State AP calculates the gradient \(g_i\) and the hessian \(h_i\) using \(y_i\) and \(\hat{y}^{(t-1)}\).
    \State AP encrypts \(g_i\) to get \(PK(g_i)\) and \(h_i\) to get \(PK(h_i)\) for each PP.
    \For{each PP} \Comment{Local Computation}
        \State Compute prefix sum on \(PK(g_i)\) and \(PK(h_i)\) across all splits and features, then forward the results to the AP.
    \EndFor
    \State AP decrypts all \(PK(g_i)\) and \(PK(h_i)\) values using \(SK\).
    \State AP evaluates each \(\mathcal{G}_i^k\) based on Equation \ref{eq:G}.
    \State AP selects the robust feature \(F_t\) and the Splitting Feature Neighborhood \(S^k_t\), then communicates these to all PPs.
    \State PP determines the best splitting point.
    \State PP forms a lookup table with entries \textbf{[record r, feature k, threshold v]}.
    \State AP chooses a robust feature randomly and establishes a response node record table with entries \textbf{[party p, record r]} for node \(t\).
\end{algorithmic}
\label{alg:split}
\vspace{-2pt}
\end{algorithm}

\subsection{Vertical Federated Unlearning}
\vspace{-5pt}
We introduce two distinct Secure Unlearning strategies: \textit{Secure Instance Unlearning} and \textit{Secure Feature Unlearning}. 

\noindent\textbf{Secure Instance Unlearning.} The Instance Unlearning process for each SecureCut tree $f_i$ begins with the active party (AP) that possesses the sample label. Initially, AP identifies the removed gradients $g_d$ and hessians $h_d$. Subsequently, the AP updates the Splitting Feature Neighborhood \(S^k_t\) in each node $t$ by $g_d$ and $h_d$. Finally, PP choose the best split point with max $\mathcal{G}$ in \(S^k_t\) and update the new instance spaces for sub-trees.

\noindent\textbf{Secure Feature Unlearning.}
The Feature Unlearning process mainly consists of three steps: First, the Active Party (AP) enumerates each tree in SecureBoost, identifying the first node whose feature is set to be unlearned. Next, the AP chooses a new robust feature and subsequently transfers ownership of the node to the feature's associated privacy preserving party, and the subtree undergoes retraining. At last, to mitigate the performance degradation caused by the unlearned feature, additional training iterations are carried out to construct new trees.

\subsection{The Explanation of Robustness}
\vspace{-5pt}
SecureCut's robustness is primarily attributed to by its Secure Binary Tree Bucketing and Robust Splitting mechanisms. Consider a binary tree, originally crafted using dataset \(D\). Even after the removal of \(D_d\), this tree retains a structure identical to one formed from \(D-D_d\). Such behavior indicates the consistency of the Secure Binary Tree Bucketing mechanism, ensuring uniform bucketing in \(M\), even if a subset \(D_d\) is removed. This mechanism assists SecureCut in creating a sturdy tree structure. Additionally, the Robust Splitting mechanism ensures that, even with the removal of a minor dataset portion, the node’s splitting feature \(k\) remains consistent.
In this cases, only the split point within the Splitting Feature Neighborhood \(S^k_t\) needs adjustment.

\begin{figure*}[tb]
    \centering
    
    \begin{minipage}{0.70\textwidth}
    
        \centering
                \captionof{table}{Comparison of utility (Acc) and forgetfulness (Wass) among different tree-based unlearning methods for Instance Unlearning when 5\% of instances are deleted from the original model. Bold values indicate the best result for each model on each dataset.}

        \vspace{-5pt}
        \begin{adjustbox}{valign=c}
            \setlength\tabcolsep{10pt}
            \resizebox{\linewidth}{!}{
            \renewcommand{\arraystretch}{0.6}
\begin{tabular}{lllllll}

\toprule
Dataset                    & Metric & Retrain & DaRE   & HedgeCut & Deltaboost & SecureCut \\ 
\midrule
\multirow{2}{*}{credit}    & Acc    & 0.874   & 0.8647 & 0.8711   & 0.8889     & \textbf{0.8903 }   \\ 
                           & Wass   & -       & 3.85   & 11.89    & 3.25       & \textbf{2.71}      \\ 
\midrule
\multirow{2}{*}{optDigits} & Acc    & 0.9661  & 0.9300 & \textbf{0.9816}   & 0.9709     & 0.9733    \\ 
                           & Wass   & -       & 4.24   & 5.78     & 1.89       & \textbf{0.97  }    \\ 
\midrule
\multirow{2}{*}{Epsilon}   & Acc    & 0.897   & 0.8883 & 0.9050   & 0.8964     & \textbf{0.9081}    \\ 
                           & Wass   & -       & 11.49  & 12.96    & 5.24       & \textbf{3.66}      \\ 
\bottomrule
\end{tabular}}
        \end{adjustbox}
        
        \label{tab:sample unlearning results}
    \end{minipage}%
    \hfill
    \begin{minipage}{0.28\textwidth}
        \centering
                \captionof{table}{The result of utility and forgetfulness in Feature Unlearning with 10\% of features are deleted.}
                \vspace{-5pt}
        \begin{adjustbox}{valign=c}

            \resizebox{\linewidth}{!}{
            \renewcommand{\arraystretch}{0.5}
\begin{tabular}{ccc}
\toprule
Dataset                    & Metric & SecureCut \\ 
\midrule
\multirow{2}{*}{credit}    & Acc    & 0.8876    \\ 
                           & Wass   & 3.57      \\ 
\midrule
\multirow{2}{*}{optDigits} & Acc    & 0.9647    \\ 
                           & Wass   & 2.68      \\ 
\midrule
\multirow{2}{*}{Epsilon}   & Acc    & 0.8983    \\ 
                           & Wass   & 5.05      \\ 
\bottomrule
\end{tabular}}
        \end{adjustbox}
        
        \label{tab:feature unlearning results}
    \end{minipage}%
    \vspace{-10pt}
\end{figure*}

\section{Experimental Results}
\vspace{-10pt}

\subsection{Experimental settings}
\vspace{-5pt}

\textbf{Datasets}. We evaluate SecureCut on three benchmark tabular datasets: \textit{Give credit} with 150k samples and 10 features, \textit{optDigits} with 5k samples and 64 features, and \textit{Epsilon} with 400k samples and 2000 features

\noindent\textbf{Baselines}. Our method is compared with several state-of-the-art unlearning methods, including: \textit{Retrain}, \textit{DaRE\cite{brophy2021machine}}, \textit{HedgeCut\cite{schelter2021hedgecut}}, and \textit{DeltaBoost\cite{wu2023deltaboost}}.

\noindent\textbf{Implementation Details.} Our method is implemented using \textit{pygbm} and \textit{mpi4py}. The model configuration consists of 100 trees. For the unlearning process, we unlearn 5\% by instances and 10\% by features.

\noindent\textbf{Metrics.} 
Firstly, we evaluate prediction accuracy on the test dataset as the \textbf{model utility} and \textit{the first Wasserstein Distance} between $M_r$ and $M_d$ as the \textbf{forgetfulness}. The first Wasserstein Distance\cite{ramdas2017wasserstein}, is often used as a metric of forgetfulness in machine unlearning\cite{tarun2023deep}:
\begin{small}
\vspace{-5pt}
\begin{equation}
    W_{1}(M_r, M_d)=\inf _{\gamma \in \Gamma(p, q)} \int_{\mathbb{R} \times \mathbb{R}}|x-y| \mathrm{d} \gamma(x, y),
    \vspace{-5pt}
\end{equation}
\end{small}
where p,q is the output distribution of  $M_r$ and $M_d$,  $\Gamma(p, q)$is the set of probability distributions whose marginal are $p$ and $q$ on the first and second factors.

\subsection{Performance Comparison and Analysis}
Tab. \ref{tab:sample unlearning results} compares SecureCut and baseline methods
in terms of the model utility and forgetfulness. 
SecureCut either achieves the best or near-best results in both forgetfulness and accuracy across all datasets. Notably, DaRE demonstrates good forgetfulness. However, its accuracy drops significantly when the datasets are not large. On the other hand, HedgeCut improves model performance, yet lacks forgetfulness. GBDT-based models outperform due to the lower variance in GBDT’s trees compared to random forests. SecureCut's superiority over all baseline models can be attributed to three main aspects:

\begin{enumerate}[leftmargin = 12pt]
\item SecureCut selects features from the Robust Features randomly for tree splitting, which reduces overfitting risks.
\item SecureCut tree structure is stable and allows rapid adjustments during unlearning.
\item SecureCut effectively filters out null values and outliers during the VFL alignment phase, enhancing model performance.
\end{enumerate}
\vspace{2pt}
As feature unlearning was not implemented in previous studies, we measured it separately. Tab. \ref{tab:feature unlearning results} shows that the accuracy of feature unlearning is well maintained, but the forgetfulness is slightly worse than instance unlearning, due to the greater structural disruption caused by feature unlearning.

\begin{figure}[tb]
    \centering
    
    \begin{subfigure}[b]{0.5\textwidth}
        \centering
        \includegraphics[width=\textwidth]{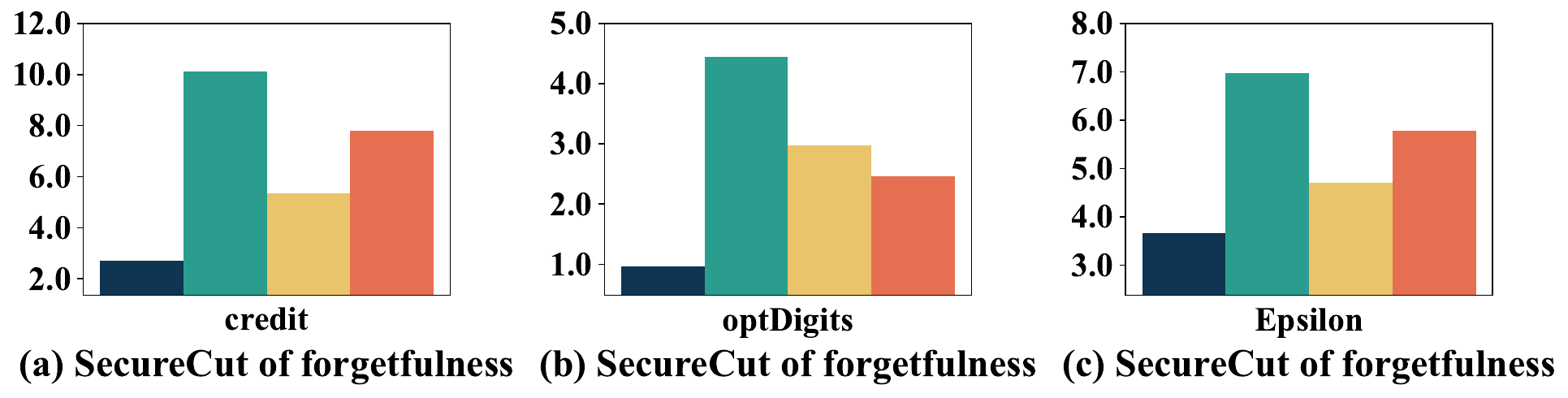}
        \label{fig:sub1}
    \end{subfigure}
    
    \vspace{-13pt} % Adjust the space between the figures as required
    
    \begin{subfigure}[b]{0.5\textwidth}
        \centering
        \includegraphics[width=\textwidth]{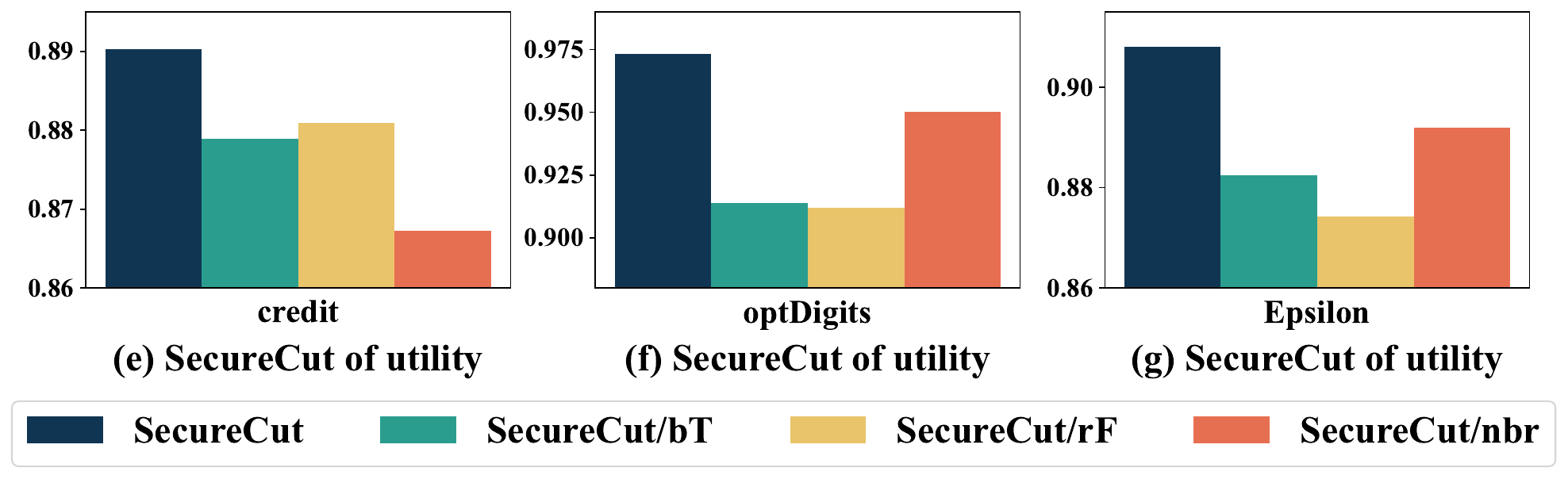}
        \label{fig:sub2}
    \end{subfigure}
    
    \vspace{-18pt}
    
    \caption{Figure (a)-(c) present the impact of different components  on forgetfulness during Unlearning task. Figure (d)-(g) present the impact of different components  on forgetfulness during Unlearning task. The Unlearning ratio is set to be 5\% }
    \label{fig:ablation}
    \vspace{5pt}
    
\end{figure}

\subsection{Ablation Study}
% \vspace{-2pt}

To further investigate the proposed SecureCut structure, we compare SecureCut with its four variants in sample unlearning: \textit{SecureCut/bT}: Trains SecureCut using normal histograms, \textit{SecureCut/rF}: Trains SecureCut utilizing the best gain feature, \textit{SecureCut/nbr}: Trains SecureCut without Splitting Feature Neighborhood.
As illustrated in Figure \ref{fig:ablation}, a)when there are fewer features, the Splitting Feature Neighborhood plays a significant role in influencing the model. This behavior arises from the tendency of the splitting points to pinpoint pronounced discontinuities. b)Yet, as the number of features augments, the Robust Features take on a more predominant role in shaping the model's performance. c)Moreover, it's imperative to retain the bucketing structure to safeguard accuracy, particularly during the unlearning process. Consequently, we ensure that each tree within SecureCut upholds a robust structural integrity.

\vspace{-10pt}
\section{Conclusion}
\vspace{-10pt}
In this paper, we propose the first machine unlearning framework, namely SecureCut, in the setting of Vertical Federated Learning. 
Specifically, we first identify two type of VFL unlearning requests, called instance unlearning and feature unlearning. Our key idea lies in our approach to create a robust tree structure, through our adoption of Secure Binary Tree Bucketing and Robust Splitting methodologies. Therefore, we build SecureCut trees in VFL setting that effectively achieves unlearning without retraining from scratch. Extensive experiments on three real-world table datasets and four tree unlearning baseline models illustrate our high utility and forgetfulness resulting from SecureCut.

\newpage

\vfill\pagebreak

\bibliographystyle{IEEEbib}
\bibliography{refs}

\end{document}